\documentclass[10pt,twocolumn,letterpaper]{article}

\usepackage{iccv}
\usepackage{times}
\usepackage{epsfig}
\usepackage{graphicx}
\usepackage{amsmath}
\usepackage{amssymb}
\usepackage[font={small}]{caption}
\usepackage{color}
\usepackage{booktabs}
\usepackage{subfig}
\usepackage{bm}
\usepackage{booktabs} 
\usepackage{multirow}
\usepackage{enumitem}
\usepackage[flushleft]{threeparttable}
\usepackage{setspace}
\usepackage{authblk}

\definecolor{my_color}{RGB}{0, 128, 255}

\usepackage[breaklinks=true,bookmarks=false]{hyperref}

\iccvfinalcopy 

\def\iccvPaperID{2498} 

\ificcvfinal\fi

\begin{document}
\hyphenpenalty=300
\tolerance=700
\hyphenation{hy-phen-a-tion}
\title{PARN: Position-Aware Relation Networks for Few-Shot Learning}


\author[]{Ziyang Wu\textsuperscript{1}}
\author[]{Yuwei Li\textsuperscript{2}}
\author[]{Lihua Guo\textsuperscript{3}}
\author[]{Kui Jia\textsuperscript{4}}

\affil[]{School of Electronic and Information Engineering, \authorcr
          South China University of Technology, Guangzhou, China\authorcr
          \tt\small \{eezywu\textsuperscript{1}, 201821010824\textsuperscript{2}\}@mail.scut.edu.cn, \{guolihua\textsuperscript{3}, kuijia\textsuperscript{4}\}@scut.edu.cn}
\renewcommand\Authands{ and }

\maketitle
\ificcvfinal\thispagestyle{empty}\fi

\begin{abstract}
    Few-shot learning presents a challenge that a classifier must quickly adapt to new classes that do not appear in the training set, given only a few labeled examples of each new class. This paper proposes a position-aware relation network (PARN) to learn a more flexible and robust metric ability for few-shot learning. Relation networks (RNs), a kind of architectures for relational reasoning, can acquire a deep metric ability for images by just being designed as a simple convolutional neural network (CNN) ~\cite{RN}. However, due to the inherent local connectivity of CNN, the CNN-based relation network (RN) can be sensitive to the spatial position relationship of semantic objects in two compared images. To address this problem, we introduce a deformable feature extractor (DFE) to extract more efficient features, and design a dual correlation attention mechanism (DCA) to deal with its inherent local connectivity. Successfully, our proposed approach extents the potential of RN to be position-aware of semantic objects by introducing only a small number of parameters. We evaluate our approach on two major benchmark datasets, \emph{\ie}, Omniglot and Mini-Imagenet, and on both of the datasets our approach achieves state-of-the-art performance with the setting of using a shallow feature extraction network. It's worth noting that our 5-way 1-shot result on Omniglot even outperforms the previous 5-way 5-shot results.
\end{abstract}
\section{Introduction} \label{Sec1}
Humans can effectively utilize prior knowledge to easily learn new concepts given just a few examples. Few-shot learning~\cite{SiameseNets,RNs1,Intro1} aims to acquire some transferable knowledge like humans, where a classifier is able to generalize to new classes when given only one or a few labeled examples of each class, \ie, one- or few-shot. In this paper, we focus on the ability of learning how to compare, namely metric-based methods. Metric-based methods~\cite{FeedNets,SiameseNets,ProtoNets,RN,MatchingNets} often consist of a feature extractor and a metric module. Given an unlabeled query image and a few labeled sample images, the feature extractor first generates embeddings for all input images, and then the metric module measures distances between the query embedding and sample embeddings to give a recognition result.

\begin{figure}[!t]
    \centering
    \includegraphics[scale=0.35]{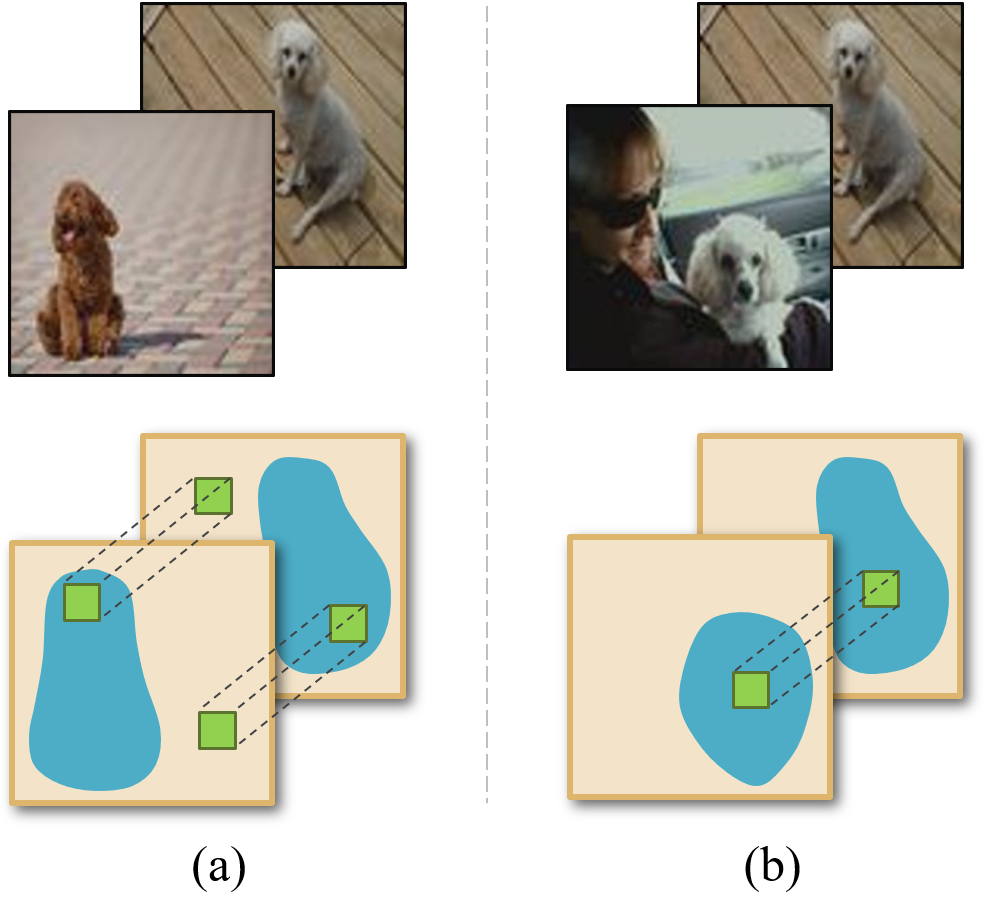}        
    \caption{Two situations where the comparison ability of RN will be limited. The top row shows the two compared images, and the bottom row shows their extracted features, where blue areas represent the response of corresponding semantic objects. (a) The convolutional kernel fails to involve the two objects. (b)  The convolutional kernel fails to involve the same fine-grained features.}
    \label{fig1}
\end{figure}

Most existing metric-based methods for few-shot learning focused on constructing a learned embedding space to better adapt to some pre-specified distance metric functions, \eg, cosine similarity~\cite{MatchingNets} or Euclidean distance~\cite{ProtoNets}. These studies expected to learn a distance metric for images, but actually only the feature embedding is learnable. As a result, the fixed but sub-optimal metric functions would limit the feature extractor to produce discriminative representations. Based on this problem, recently Sung~\etal~\cite{RN} introduced a relation network, which was designed as a simple CNN, to make the metric learnable and flexible in a data-driven way (in this paper we denote the \emph{simply CNN-based relation network} as RN), and they achieved impressive performance in few-shot learning. However, according to our analysis, the comparison ability of RN is still limited due to its inherent local connectivity.

As we all know, convolutional operations naturally have the translation invariance to extract features from images, meaning that higher responses of extracted features mainly locate in positions corresponding to the semantic objects~\cite{VisualCNNs}. There are two situations: (i) two semantic objects of images are in totally different spatial positions, as shown in Figure~\ref{fig1}(a); (ii) they are in close spatial positions while their fine-grained features do not, as shown in Figure~\ref{fig1}(b). We note that these two situations commonly occur in the datasets, especially the situation (ii), which should not be overlooked. For these two situations, Sung~\etal~\cite{RN} simply concatenated two compared features together and used RN to learn their relationship. However, we argue that the comparison ability of RN is inherently constrained due to its local receptive fields. In situation (i), as shown in Figure~\ref{fig1}(a), each convolution step can only involve a same local spatial region, which rarely contains two objects at the same time. In situation (ii), even if the convolutional kernel involves two objects simultaneously, it may also fail to involve their related fine-grained semantic features, \eg, in Figure~\ref{fig1}(b) it involves body features of the sample and head features of the query, which is not optimal and reasonable as a comparison operation. These two situations motivate us to promote RN aware of objects and fine-grained features in different positions.

In this paper, we propose a position-aware relation network (PARN), where the convolution operator can overcome its local connectivity to be position-aware of related semantic objects and fine-grained features in images. Compared with RN~\cite{RN}, our proposed model provides a more efficient feature extractor and a more robust deep metric network, which enhances the generalization capability of the model to deal with the above two situations. The overall framework is shown in Figure~\ref{fig2}. Our main contributions are as follows:
\begin{itemize}
\setlength{\topsep}{0pt}
\setlength{\itemsep}{0pt}
\setlength{\parsep}{0pt}
\setlength{\parskip}{0pt}
\item During the feature extraction phase, we introduce the deformable feature extractor (DFE) to extract more efficient features, which contain fewer low-response or unrelated semantic features, for effectively alleviating the problem in the situation (i).
\item Our another important contribution is that we further exploit the potential of RN to be position-aware to learn a more robust and general metric ability. During the comparison phase, we propose a dual correlation attention mechanism (DCA) that utilizes position-wise relationships of two compared features to capture their global information, and then densely aggregate the captured information into each position of outputs. In this way, the subsequent convolutional layer can sense related fine-grained features in all positions, and adaptively compare them despite of the local connectivity.
\item With the setting of using a shallow feature extraction network, our method achieves state-of-the-art results with a comparable margin on two major benchmarks, \ie, Omniglot and Mini-Imagenet. It's worth noting that our 5-way 1-shot result on Omniglot even outperforms the previous 5-way 5-shot results.
\end{itemize}

\section{Related Work}

Recent methods for few-shot learning usually adopted the \emph{episode}-based strategy~\cite{MatchingNets} to learn meta-knowledge from a set of episodes, where each episode/task/mini-batch contains $C$ classes and $K$ samples of each class, \ie, $C$-way $K$-shot. The acquired meta-knowledge could enable the model to adapt to new tasks that contain unseen classes with only a few samples. According to the variety of meta-knowledge, recent methods could be summarized into the three categories, \ie, optimization-based (learning to optimize the model quickly)~\cite{MAML,MetaLSTM,MetaGAN,MetaFGNet}, memory-based (learning to accumulate and generalize experience)~\cite{MMNets,MetaNets,MANN} and metric-based (learning a general metric)~\cite{FeedNets,SiameseNets,ProtoNets,RN,MatchingNets} methods.

Briefly, optimization-based methods usually associated with the concept of meta-learning/learning to learn~\cite{MetaLearning2,MetaLearning1}, \eg, learning a meta-optimizer~\cite{MetaLSTM} or taking some wise optimization strategies~\cite{MAML,MetaGAN,MetaFGNet}, to better and faster update the model for new tasks. Memory-based methods generally introduced memory components to accumulate experience when learning old tasks and generalize them when performing new tasks~\cite{MMNets,MetaNets,MANN}. Our experimental results show that our method outperforms them without the need for updating the model for new tasks or introducing complicated memory structure.

Metric-based methods, where our approach belongs to, can perform new tasks in a feed-forward manner, which often consist of a feature extractor and a metric module. The feature extractor first generates embeddings for the unlabeled query image and a few labeled sample images, and then the recognition result is given by measuring distances between the query embedding and sample embeddings in the metric module. Earlier works~\cite{FeedNets,SiameseNets,ProtoNets,MatchingNets} mostly focused on designing embedding methods or some well-performed but fixed metric mechanism. For example, Bertinetto~\etal~\cite{FeedNets} designed a task-adaptive feature extractor for new tasks by utilizing a trained network to predict parameters. And Vinyals~\etal~\cite{MatchingNets} proposed a learnable attention mechanism by introducing LSTM to calculate fully context embeddings (FCE), and applying softmax over the cosine similarity in the embedding space, which developed the idea of a fully differentiable neural neighbors algorithm. Yet their approach was somewhat complicated. Snell~\etal~\cite{ProtoNets} then further exceeded them with prototypical networks by simply learning an embedding space, where prototypical representations of classes could be obtained by directly calculating the mean of samples, and they used Bregman divergences~\cite{Diver} to measure distance, which outperforms the cosine similarity used in \cite{MatchingNets}.

In the above metric-based methods, embeddings would be limited to produce discriminative representations in order to meet the fixed but sub-optimal metric methods. Some approaches~\cite{Mashi1,Mashi2} tried to adopt the Mahalanobis metric, while still inadequate in the high-dimensional embedding space. To solve this problem, Sung~\etal~\cite{RN} introduced relation networks (RNs) for few-shot learning, which are a kind of architectures for relational reasoning and successfully applied in visual question answering tasks~\cite{RNs3,RNs1,RNs2}. They achieved impressive performance by designing a simply CNN-based relation network (RN) to develop a learnable non-linear metric module, which is simple but flexible enough for the embedding network. However, due to the local connectivity of CNN, RN would be sensitive to the spatial position relationship of compared objects. Therefore, we further exploit the potential of RN to learn a more robust metric ability, which avoids this problem.

\section{Approach}

In this section, we give the details of the proposed position-aware relation network (PARN) for few-shot learning. At first, we will present the overall framework of PARN. Then we will introduce our deformable feature extractor (DFE) which could extract more efficient features. At last, to promote RN position-aware of fine-grained features in images, we propose a dual correlation attention mechanism (DCA).

\subsection{Overall}

\begin{figure}[tb]
    \centering
    \includegraphics[scale=0.4]{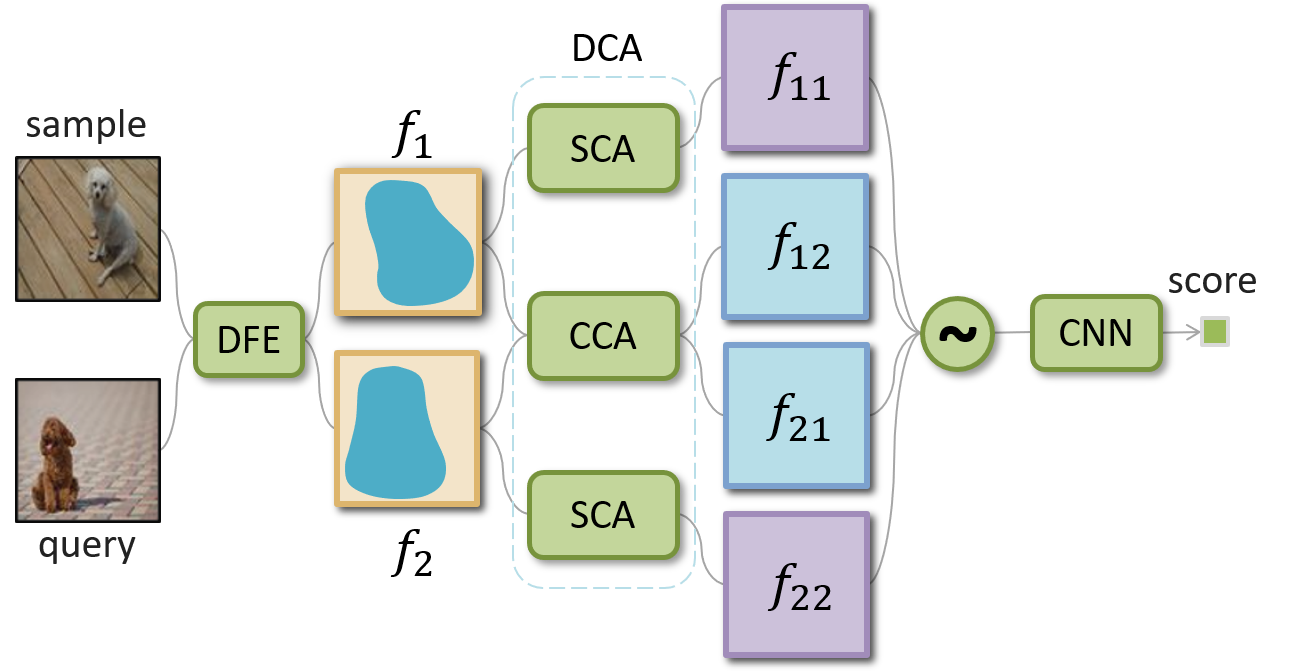}        
    \caption{Overview of our proposed PARN for few-shot learning. DFE is the deformable feature extractor. DCA is the dual correlation attention module, which consists of a cross-correlation attention module (CCA) and a self-correlation attention module (SCA). The two SCA blocks are a shared module. The symbol `$\bm{\sim}$' represents a concatenating operation.}
    \label{fig2}
\end{figure}

The network architecture is given in Figure~\ref{fig2}. At first, a sample and a query image are fed into a feature extraction network, which is designed as a DFE. With DFE, extracted features $\bm{f_1}$ and $\bm{f_2}$ can be more focused on the semantic objects, which is beneficial to improve the subsequent comparison efficiency and precision.

Then, in order to make a robust comparison between $\bm{f_1}$ and $\bm{f_2}$, we apply the dual correlation attention module (DCA) over them, so that each position of the output feature map $\bm{f_{mn}}(\bm{m},\bm{n}\in\{1,2\})$ contains global cross- or self-correlation information, where $\bm{f_{mn}}$ means that each position of $\bm{f_m}$ attends to all positions of $\bm{f_n}$. In this way, even if the subsequent convolution operations are locally connected, each convolution step can adaptively sense related fine-grained semantic features in all positions.

Finally, we concatenate the above output features $\bm{f_{mn}}(\bm{m},\bm{n}\in\{1,2\})$, and feed them into a standard CNN to learn the relation score.

\subsection{Deformable Feature Extractor} \label{DFE}

\begin{figure}[tb]
    \centering
    \includegraphics[scale=0.32]{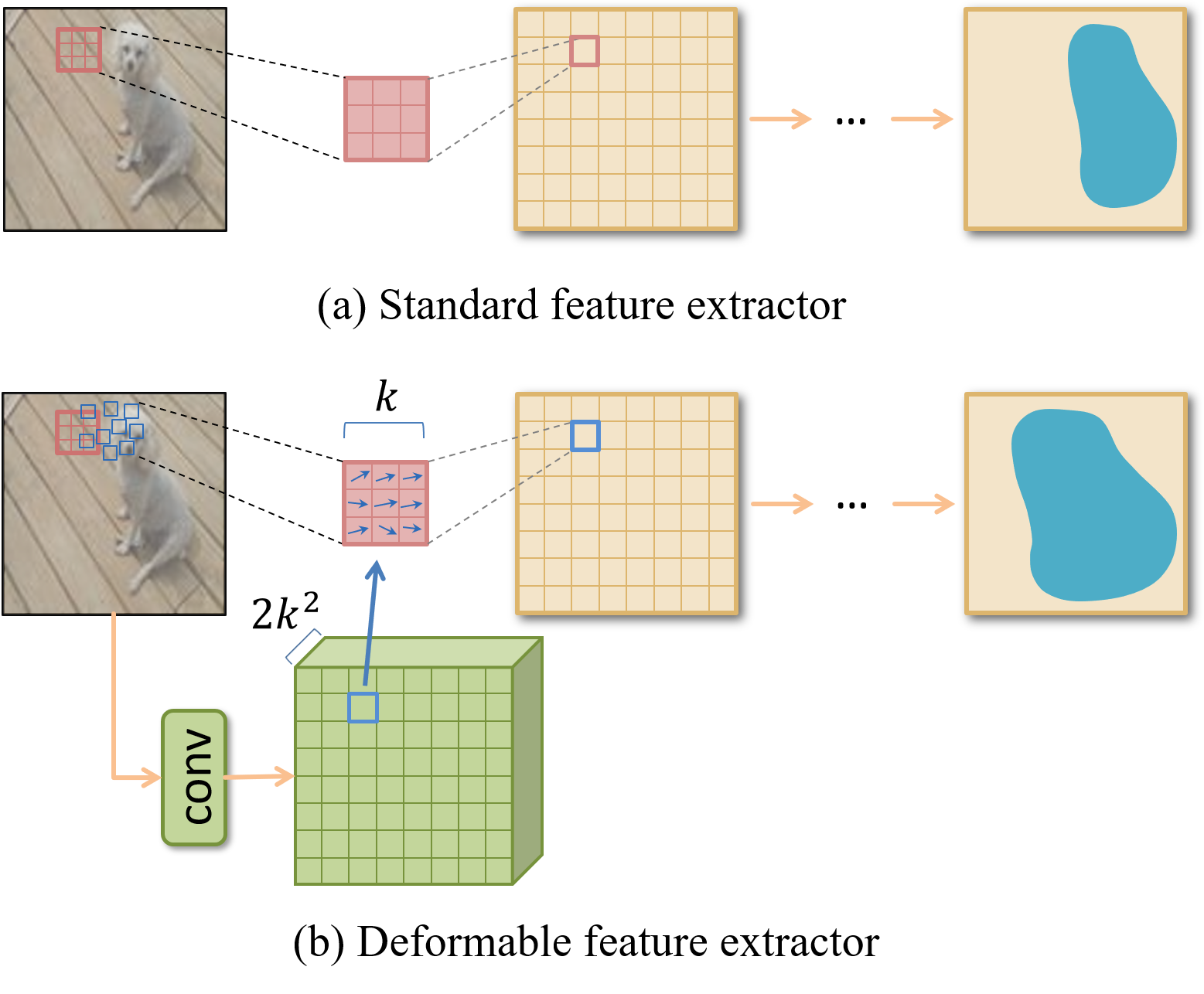}        
    \caption{Two feature extractors. Feature maps are shown in spatial shapes. Blue areas on output features represent the response of corresponding semantic objects.}
    \label{fig3}
\end{figure}

Figure~\ref{fig3}(a) shows a standard feature extractor (SFE). Due to the translation invariance of convolutional operations, the output feature extracted by SFE would only present high responses in spatial positions corresponding to the object. Other positions are low-response or unrelated features that may induce the metric module to perform some redundant comparison operations on them, which affects the efficiency of the comparison. In the worst scenario like Figure~\ref{fig1}(a), it is difficult to accurately compare the two objects.

Inspired by the idea of deformable convolutional networks~\cite{DCN,STN} for object detection tasks, we try to deploy deformable convolutional layers for the feature extraction network to extract more efficient features that contain fewer low-response or unrelated semantic features. As shown in Figure~\ref{fig3}(b), the convolutional kernel of a deformable convolutional layer is not a regular $k\times k$ grid, but $k^2$ parameters with 2D offsets. Each parameter $\bm{w_i}(0\leq i\leq k^2)$ of the kernel should take an offset coordinate $(\Delta x, \Delta y)$, transforming the original operation from $\bm{w_i}*\bm{f_{(x,y)}}$ to $\bm{w_i}*\bm{f_{(x+\Delta x,y+\Delta y)}}$, where $\bm{f_{(x,y)}}$ refers to a spatial point at the coordinate $(x,y)$ of $\bm{f}$. In our work, the offsets are learned by applying a convolutional layer over the input feature map following Dai~\etal~\cite{DCN}. And the offsets map has the same spatial resolution as the output map, while its channel dimension is $2k^2$, since for every spatial position of the output map there are $k\times k\times 2=2k^2$ offset scalars.

Comparing the features extracted by SFE and DFE in Figure~\ref{fig3}(a)(b), we can learn that DFE can filter out unrelated information to some extent, and extract a more efficient feature, which is expected to improve the subsequent comparison efficiency and performance.

\subsection{Dual Correlation Attention Module} \label{DCA}

Despite of more efficient features, as mentioned in Section~\ref{Sec1}, if we just use convolutional operations to implement the subsequent comparison procedure, the comparison ability is still limited, since it is somewhat difficult to involve related fine-grained semantic features of the two images at each convolution step. To deal with this problem, one immediate idea is to use a larger receptive field by enlarging the size of the convolutional kernel, or stacking several convolutional layers. However, with more parameters and deeper layers, the model will fall into overfitting problems more easily.

Inspired by the non-local networks~\cite{Non-local} that captures long-term dependencies for video classification task, we propose a dual correlation attention mechanism (DCA) for the two-input deep relation network. The proposed attention mechanism uses just a small number of parameters to capture relationships between any two positions of features, regardless of their spatial distance, and then utilizes the captured position-wise relationships to aggregate global information at each spatial position of outputs. In this way, even if the subsequent convolutional kernel is small, each convolution step can involve global information of the two input features, and adaptively perform the comparison on them.
\begin{figure*}[!t]
    \centering
    \includegraphics[scale=0.32]{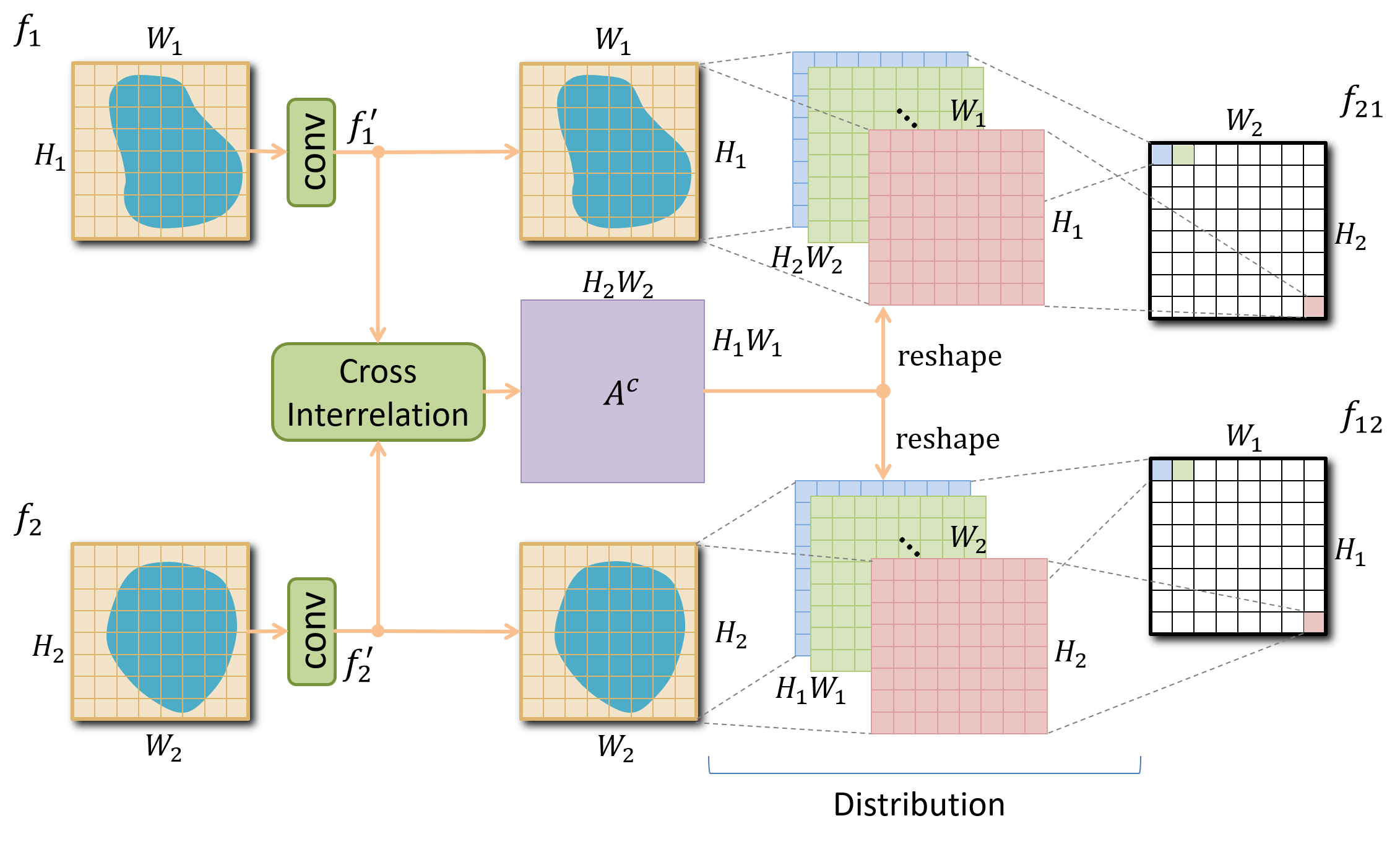}        
    \caption{The cross-correlation attention module (CCA). Feature maps are shown in spatial shapes. Weights of the two $1\times 1$ convolutional layers are shared. The cross-correlation attention map $\bm{A^c}$ contains all the position-wise correlationships of the two inputs. During the distribution operation, $\bm{A^c}$ will be reshaped into shapes corresponding to the spatial shape of $\bm{f_1}$ (or $\bm{f_2}$). Each sub-map of $\bm{A^c}$ is then performed dot-product with $\bm{f^{'}_1}$ (or $\bm{f^{'}_2}$) to aggregate cross-global information into each spatial position of the output $\bm{f_{21}}$ (or $\bm{f_{12}}$).}
    \label{fig4}
\end{figure*}

\begin{figure*}[!t]
    \centering
    \includegraphics[scale=0.32]{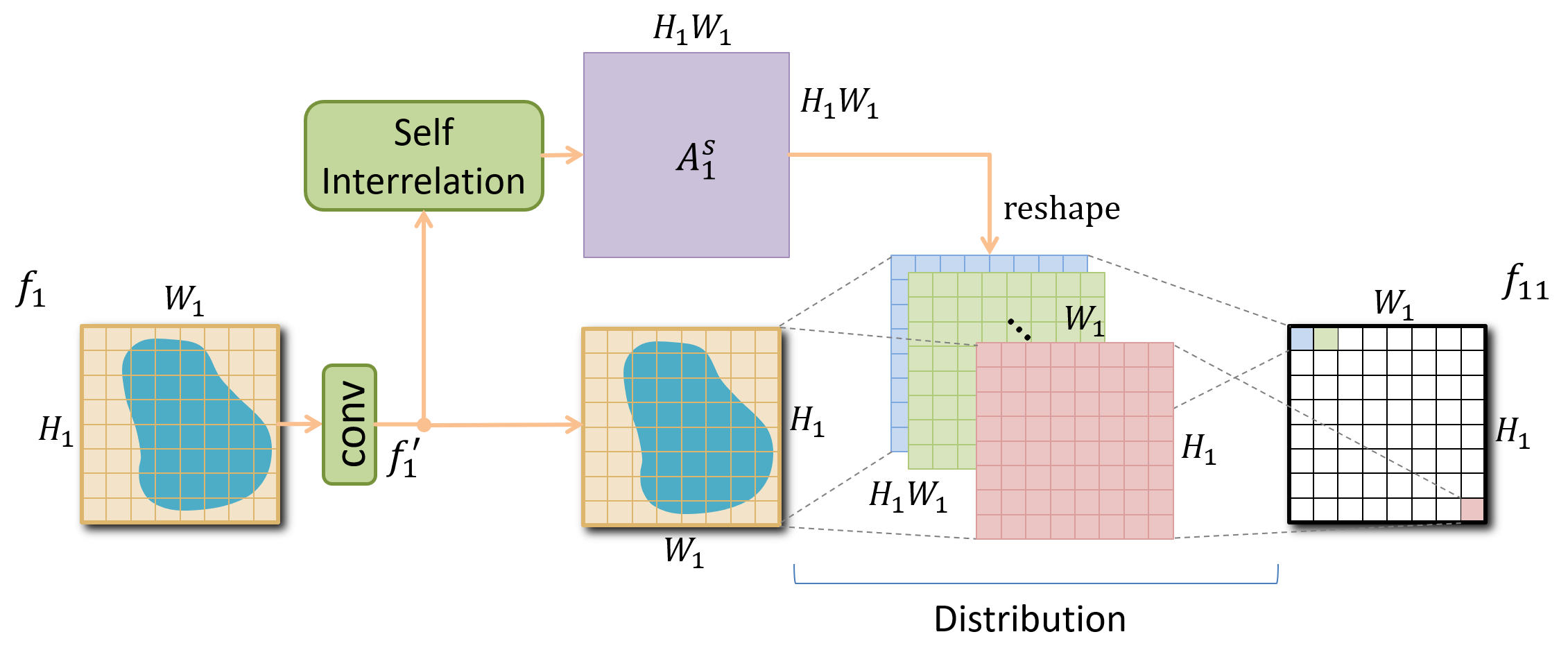}        
    \caption{The self-correlation attention module (SCA). Feature maps are shown in spatial shapes. Weights of the $1\times 1$ convolutional layer are shared with that in CCA. The self-correlation attention map $\bm{A^s_1}$ contains all the position-wise relationships in $\bm{f_1}$. Each sub-map of $\bm{A^s_1}$ is then performed dot-product with $\bm{f^{'}_1}$ to aggregate global information into each spatial position of the output $\bm{f_{11}}$.}
    \label{fig5}
\end{figure*}
As shown in Figure~\ref{fig2}, the proposed DCA consists of a cross-correlation attention module (CCA) and a self-correlation attention module (SCA), where CCA calculates $\bm{f_{12}}$ (or $\bm{f_{21}}$) by attending every spatial position of $\bm{f_1}$ (or $\bm{f_2}$) to the global information of $\bm{f_2}$ (or $\bm{f_1}$), and SCA calculates $\bm{f_{11}}$ (or $\bm{f_{22}}$) by attending every spatial position of $\bm{f_1}$ (or $\bm{f_2}$) to the global information of its own. We will give their details respectively below.




\begin{spacing}{0.4}
    $\\$
\end{spacing}
\textbf{Cross-correlation attention module}\quad As shown in Figure~\ref{fig4}, given two extracted features $\bm{f_1}\in\mathbb{R}^{C\times H_1\times W_1}$ and $\bm{f_2}\in\mathbb{R}^{C\times H_2\times W_2}$\footnote{Actually $H_1$ and $W_1$ are equal to $H_2$ and $W_2$. Here we denote them as different notations for clear explanation.}​, CCA first applies two shared $1\times1$ convolutional layers over them respectively to make a embedding over the channel dimension, and then generates two feature maps $\bm{f^{'}_1}\in\mathbb{R}^{C^{'}\times H_1\times W_1}$ and $\bm{f^{'}_2}\in\mathbb{R}^{C^{'}\times H_2\times W_2}$, where $C^{'}$ is less than $C$. We reshape them into $\bm{f^{'}_1}\in\mathbb{R}^{H_1W_1\times C^{'}}$ and $\bm{f^{'}_2}\in\mathbb{R}^{H_2W_2\times C^{'}}$.
Then we apply a cross-interrelation operation $\bm{g(f^{'}_1, f^{'}_2)}$ to calculate their relationships of any two positions into the cross-attention map $\bm{A^c}$. From the spatial position $\bm{i}$ of $\bm{f^{'}_1}$ and $\bm{j}$ of $\bm{f^{'}_2}$, we can respectively get two spatial points/vectors $\{\bm{f^{'}_{1i}}, \bm{f^{'}_{2j}}\}\in\mathbb{R}^{C^{'}}$, where $\bm{i}\in\{1,...,H_1W_1\}, \bm{j}\in\{1,...,H_2W_2\}$. The pointwise calculation of $\bm{g(f^{'}_1, f^{'}_2)}$ is denoted as $\bm{g_{ij}(f^{'}_{1i}, f^{'}_{2j})}$, \ie, $\bm{g_{ij}}$ computes the value of $\bm{A^c_{ij}}$, which indicates the relationship between $\bm{f^{'}_{1i}}$ and $\bm{f^{'}_{2j}}$. Here we choose the cosine similarity function for $\bm{g_{ij}}$ to calculate their relationships, then $\bm{A^c_{ij}}$ can be computed as follows:
\begin{equation}\label{eq1}
    \bm{A^c_{ij}}=\bm{g_{ij}(f^{'}_{1i}, f^{'}_{2j})}=\bm{\overline{f}^{'}_{1i}\overline{f}^{'T}_{2j}}
\end{equation}
where $\bm{\overline{f}^{'}_{1i}}=\bm{\frac{f^{'}_{1i}}{\|f^{'}_{1i}\|}}$ and $\bm{\overline{f}^{'}_{2j}}=\bm{\frac{f^{'}_{2j}}{\|f^{'}_{2j}\|}}$ are the $l_2$-normalized vectors. We denote $\bm{\overline{f}^{'}_{1}}=[\bm{\overline{f}^{'}_{1i}}]\in\mathbb{R}^{H_1W_1\times C^{'}}$ and $\bm{\overline{f}^{'}_{2}}=[\bm{\overline{f}^{'}_{2j}}]\in\mathbb{R}^{H_2W_2\times C^{'}}$, meaning that $\bm{\overline{f}^{'}_{1}}$ and $\bm{\overline{f}^{'}_{2}}$ are obtained by performing $l_2$-normalization over $\bm{f^{'}_1}$ and $\bm{f^{'}_2}$ respectively along their channel dimension. Then Eq.~\eqref{eq1} can be rewritten in matrix form:
\begin{equation}\label{eq2}
    \bm{A^c}=\bm{g(f^{'}_{1}, f^{'}_{2})}=\bm{\overline{f}^{'}_{1}\overline{f}^{'T}_{2}}
\end{equation}
where $\bm{A^c}\in\mathbb{R}^{H_1W_1\times H_2W_2}$ contains all the correlationships between every spatial position of $\bm{f^{'}_1}$ and $\bm{f^{'}_2}$.

After obtaining the cross-attention map $\bm{A^c}$, as shown in Figure~\ref{fig4}, the next step is the distribution operation that performs dot-product between each sub-map of $\bm{A^c}$ with $\bm{f^{'}_1}$ and $\bm{f^{'}_2}$ respectively. We perform the distribution as follows:
\begin{equation}\label{eq3}
    \left\{
        \begin{array}{lr}
            \bm{f_{21}}=\bm{{A^c}^Tf^{'}_1}&  \\
            \specialrule{0em}{0.5ex}{0.5ex}
            \bm{f_{12}}=\bm{A^cf^{'}_2}&
        \end{array}
    \right. 
\end{equation}
where $\bm{f_{mn}}$ means that $\bm{f_m}$ attends to the global information of $\bm{f_n}$ $(\bm{m},\bm{n}\in\{1,2\}, \bm{m}\not=\bm{n})$. Specifically, we can learn from Figure~\ref{fig4} that the output feature $\bm{f_{21}}$ captures the global information of $\bm{f_1}$ into each its spatial position, and so does $\bm{f_{12}}$ to $\bm{f_2}$. In this way, the subsequent convolutional layer can sense all the positions, and compare them even with a small convolutional kernel. At last $\bm{f_{21}}$ and $\bm{f_{12}}$ will be reshaped into $\bm{f_{21}}\in\mathbb{R}^{C^{'}\times H_2\times W_2}$ and $\bm{f_{12}}\in\mathbb{R}^{C^{'}\times H_1\times W_1}$ respectively, and then pass through a $1\times1$ convolutional layer to increase the channel dimension to $C$.





\begin{spacing}{0.4}
    $\\$
\end{spacing}
\textbf{Self-correlation attention module}\quad As shown in Figure~\ref{fig5}, SCA is similar to CCA in Figure~\ref{fig4}, except that the self-interrelation operation in SCA accept only one input to generate a self-attention map $\bm{A^s}$, which is actually the case when two inputs of the cross-interrelation operation are the same in our implementation. Besides, the weights of the two $1\times 1$ convolutional layers in SCA are shared with that in CCA. Therefore, referring to Eq.~\eqref{eq2}\eqref{eq3}, given the input feature $\bm{f_1}$, we can also get the output $\bm{f_{11}}$:
\begin{equation}\label{eq4}
    \bm{A^s_1}=\bm{g(f^{'}_{1}, f^{'}_{1})}=\bm{\overline{f}^{'}_{1}\overline{f}^{'T}_{1}}
\end{equation}
\begin{equation}\label{eq5}
    \bm{f_{11}}=\bm{{A^s_1}^Tf^{'}_1}
\end{equation}
where $\bm{f_{11}}$ means $\bm{f_1}$ attends to itself, and captures the global information to aggregate into each its spatial position. By inputting $\bm{f_2}$ and performing the same operations, we can also get $\bm{A^s_2}$ and $\bm{f_{22}}$. The next step for $\bm{f_{11}}$ and $\bm{f_{22}}$ is the same as for $\bm{f_{12}}$ and $\bm{f_{21}}$.

Then the computations of DCA are completed, where all the introducing parameters are only one shared $1\times 1$ convolutional layer for embedding input features and another shared $1\times 1$ convolutional layer for increasing the channel dimension. After that, we concatenate these four globally related features $\bm{f_{mn}}(\bm{m},\bm{n}\in\{1,2\})$\footnote{In our experiments we also concatenate the two input features.}​ and pass through a CNN to learn the final relation score.

\section{Experiments}

In this section, we first introduce two benchmark datasets and implementation details. Then we conduct a series of ablation studies to analyze the effectiveness of our proposed model. Finally we compare our proposed model with previous state-of-the-art methods on these two datasets.

\subsection{Datasets}

\textbf{Omniglot}~\cite{Omniglot} is a common benchmark for few-shot learning, which contains 1,623 different handwritten characters/classes from 50 different alphabets, and each class has a maximum of 20 samples of size $28\times 28$. We follow the standard splits \cite{ProtoNets,RN,MatchingNets} that there are 1,200 classes for meta-training and 423 classes for meta-testing. In addition, we follow \cite{MANN,ProtoNets,MatchingNets} to augment the dataset with random rotations by multiples of 90 degrees during training.

\textbf{Mini-Imagenet}~\cite{MatchingNets} is a subset of Imagenet, consisting of 100 classes, each of which contains 600 images of size $84\times 84$. We follow \cite{MAML,MetaLSTM,ProtoNets,RN,MatchingNets} in the exactly same way to split the dataset, \ie, 64 classes for meta-training, 16 classes for meta-validation and 20 classes for meta-testing.

\subsection{Implementation Details}

\textbf{Network architectures}\quad Following the previous works~\cite{ProtoNets,RN,MatchingNets}, our basic feature extraction network, the standard feature extractor (SFE), consists of 4 convolutional modules, each of which contains a 64-filter of $3\times 3$ convolutions, followed by batch normalization~\cite{BN} and ReLU nonlinearity. Besides, we apply $2\times 2$ max-pooling in the last two layers. As for the basic relation network (RN), we follow the same architecture in \cite{RN}, namely two convolutional modules with 64-filter, followed by two fully connected layers, and the final output is mapped into 0-1 as the relation score through a sigmoid function.

\begin{spacing}{0.4}
    $\\$
\end{spacing}
\textbf{Training and testing details}\quad We implement all the experiments in Pytorch with a GeForce GTX 1080 Ti GPU. We use Adam~\cite{Adam} to optimize the network end-to-end, starting with a learning rate of 0.001 and reducing it by a factor of 10 when the validation accuracy stopped improving. We use the mean square error (MSE) loss to train the network as a regression task, where the label is 1 when the two input categories are the same, otherwise 0. No regularization techniques such as dropout or $l_2$ regularisation are applied during training. We follow Sung~\etal~\cite{RN} to arrange the number of sample and query images for the 1-shot and 5-shot tasks. The classification result is given by the category with the highest score.

\subsection{Ablation Study}

In this subsection, we do some ablation experiments on Mini-Imagenet to examine the effectiveness of DFE and DCA.

\begin{spacing}{0.4}
    $\\$
\end{spacing}
\begin{table}[b]
    \small
    \centering
    \setlength{\tabcolsep}{1.4mm}{
    \begin{tabular}{ccccc}
    \toprule
    Model&5-way 1-shot&5-way 5-shot&params&depth\\
    \midrule
    SFE-4&51.64 $\pm$ 0.83\%&66.08 $\pm$ 0.69\%&0.424M&4\\
    SFE-6&51.74 $\pm$ 0.84\%&67.13 $\pm$ 0.67\%&\textbf{0.498M}&\textbf{6}\\
    DFE-4&\textbf{52.07 $\pm$ 0.82\%}&\textbf{67.53 $\pm$ 0.67\%}&0.445M&4\\
    \bottomrule
    \end{tabular} }
    \caption{The ablation study of DFE on Mini-Imagenet. Results are obtained by averaging over 600 test episodes with 95\% confidence intervals.}
    \label{table1}
\end{table}


\begin{figure}[b]
    \small
    \centering
    \includegraphics[scale=0.5]{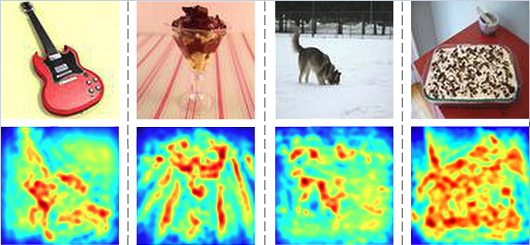}        
    \caption{Visualization of the effective receptive fields (ERF)~\cite{ERF} of DFE. DFE can filter out some useless information, such as the background.}
    \label{fig6}
\end{figure}
\textbf{Deformable feature extractor}\quad In Section~\ref{DFE}, we propose DFE to extract more efficient features, which is expected to improve the subsequent comparison efficiency and precision. To validate the expectation, we observe the results of using SFE with 4 convolutional layers (SFE-4) or DFE with 4 convolutional layers (DFE-4) to extract features for the subsequent comparison. The structures of SFE-4 and DFE-4 are the same, except that the last two convolutional layers of DFE-4 are deformable convolutional layers. To eliminate the influence of extra parameters introduced by DFE-4, we set up SFE with 6 convolutional layers (SFE-6) for comparison. In this ablation experiment, we just use RN without DCA as the metric network. As we find that the learning of deformable convolutional layers tends to be unstable at the begining, we initialize the parameters of the convolutional layer that learns offsets to be 0 and start training them after about 10000 episodes of warm-up.

The results are shown in Table~\ref{table1}. It can be seen that by using DFE, the accuracies are improved from 51.64\% to 52.07\% in the 5-way 1-shot task and 66.08\% to 67.53\% in the 5-way 5-shot task, and slightly better than SFE-6 that holds more parameters, which indicates the effectiveness of DFE. In Figure~\ref{fig6}, we further visualize the effective receptive field (ERF)~\cite{ERF} of DFE on the input images. The visualization shows that the learned offsets in the deformable convolutional layers can potentially adapt to the image object, meaning that DFE can filter out some useless information to extract more efficient features, which helps the subsequent comparison procedure. Note that ERF does not represent the response of extracted features, but just represents the effective area in the receptive field, that is, the network is watching at these places. So it is acceptable if DFE just filters out some background information, but does not exactly focus on desirable objects.

\begin{spacing}{0.4}
    $\\$
\end{spacing}

\textbf{Dual correlation attention mechanism}\quad In this ablation experiment, we take SFE as the feature extractor and RN as the basic metric network. So when no proposed attention module is used, the overall network is our reimplementation of RN in \cite{RN}. To verify our proposed DCA, we conduct experiments on whether RN is applied with CCA, SCA or their combination DCA. For fair comparison, a simple $1\times 1$ convolutional layer will be added before RN as the baseline of the proposed attention modules.

The results are shown in Table~\ref{table2}. We can see that in 1-shot and 5-shot tasks, the proposed CCA and SCA both improve the performance. Especially when combining the two modules as DCA, the accuracies increase to 54.36\% in the 1-shot task and 70.50\% in the 5-shot task, which outperforms the baseline by a clear margin. Besides, we find that during training the network converges much faster with DCA, indicating that DCA successfully allows RN to perceive related semantic features in different positions, and makes it easier to learn to compare.

\begin{figure}[tb]
    \centering
    \includegraphics[scale=0.5]{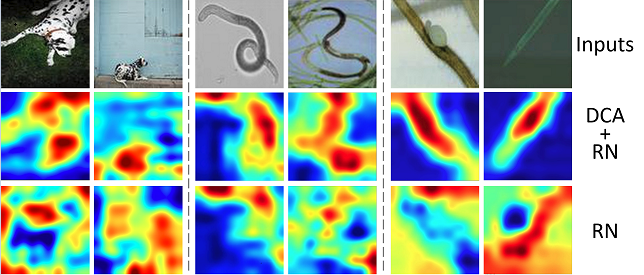}        
    \caption{Three Visualization examples of the Gradient-weighted Class Activation Mapping (Grad-CAM)~\cite{GradCAM} on two input images for RN with or without DCA. With DCA, RN successfully compares related semantic features of two images in different positions, while without DCA it fails to do it.}
    \label{fig7}
\end{figure}

To more intuitively observe the effectiveness of DCA, we use the gradient-weighted class activation mapping (Grad-CAM) introduced in \cite{GradCAM} to visualize the output result activations on the two compared images. As shown in Figure~\ref{fig7}, when the related fine-grained semantic features of two objects are in different positions, RN fails to compare them without our proposed DCA, while with DCA it can successfully do it. In other words, with the proposed DCA, RN become more robust and general to learn metrics.



It is worthy to notice that CCA works much better than SCA as shown in Table~\ref{table2}. We analyze that the main reason may attribute to the certain ability of preliminary comparison of CCA, while SCA does not have it. As mentioned in Section~\ref{DCA}, the cross-attention map $\bm{A^c}$ of CCA is calculated by the cross-interrelation operation $\bm{g(f_1,f_2)}$, which is actually implemented by a similarity function. Therefore, when two input features come from different categories, most values of $\bm{A^c}$ will tend to be smaller. Then in Eq.~\eqref{eq3}, since $\bm{f^{'}_1}$ and $\bm{f^{'}_2}$ are relatively stable after the BN~\cite{BN} layer of SFE, we can infer that the response of $\bm{f_{12}}$ and $\bm{f_{21}}$ will tend to be lower due to the small $\bm{A^c}$. In other words, inputs of different categories lead to small outputs. While the situation is opposite when $\bm{f_1}$ and \bm{$f_2}$ come from the same category. So we can learn that the outputs of CCA have preliminarily represented the relationship between the two inputs, which can help the subsequent RN to make further comparisons.

\begin{table}[tb]
    \small
    \centering
    \setlength{\tabcolsep}{1.4mm}{
    \begin{tabular}{ccc}
    \toprule
    \multirow{2}{*}{Method} &\multicolumn{2}{c}{\textbf{5-way Acc.}}\\
     \cmidrule(l){2-3}
      &\begin{tabular}[c]{@{}c@{}}1-shot\end{tabular}&\begin{tabular}[c]{@{}c@{}}5-shot\end{tabular} \\
      \midrule
    RN&51.64 $\pm$ 0.83\%&66.08 $\pm$ 0.69\%\\
    baseline&51.29 $\pm$ 0.82\%&66.00 $\pm$ 0.70\% \\
    SCA&52.64 $\pm$ 0.91\%&67.14 $\pm$ 0.70\% \\
    CCA&53.88 $\pm$ 0.87\%&69.49 $\pm$ 0.69\% \\
    CCA\&SCA&\textbf{54.36 $\pm$ 0.84\%}&\textbf{70.50 $\pm$ 0.64\%} \\
    \bottomrule
    \end{tabular} }
    \caption{The ablation study of DCA on Mini-Imagenet. The baseline is a $1\times 1$ convolutional layer with RN. The combination of SCA and CCA is the proposed DCA. Results are obtained by averaging over 600 test episodes with 95\% confidence intervals.}
    \label{table2}
\end{table}

\begin{table*}[!tb]
    \small
    \centering
    \setlength{\tabcolsep}{1.4mm}{
    \begin{tabular}{ccccc}
    \toprule
    \multirow{2}{*}{Method} &\multicolumn{2}{c}{\textbf{5-way Acc.}}&\multicolumn{2}{c}{\textbf{20-way Acc.}}\\
    \cmidrule(l){2-3}\cmidrule(l){4-5}
        &\begin{tabular}[c]{@{}c@{}}1-shot\end{tabular}&\begin{tabular}[c]{@{}c@{}}5-shot\end{tabular}&\begin{tabular}[c]{@{}c@{}}1-shot\end{tabular}&\begin{tabular}[c]{@{}c@{}}5-shot\end{tabular} \\
    \midrule
    MANN~\cite{MANN}&82.8\%&94.9\%&-&- \\
    Matching Nets~\cite{MatchingNets}&98.1\%&98.9\%&93.8\%&98.5\% \\
    Siamese Nets~\cite{SiameseNets}&98.4\%&99.6\%&95.0\%&98.6\% \\
    Meta Nets~\cite{MetaNets}&98.95\%&-&97.0\%&- \\
    Proto Nets~\cite{ProtoNets}&97.4\%&99.3\%&95.4\%&98.7\% \\
    MAML~\cite{MAML}&98.7 $\pm$ 0.4\%&99.9 $\pm$ 0.1\%&95.8 $\pm$ 0.3\%&98.9 $\pm$ 0.2\% \\
    MMNet~\cite{MMNets}&99.28 $\pm$ 0.08\%&99.77 $\pm$ 0.04\%&97.16 $\pm$ 0.10\%&98.93 $\pm$ 0.05\% \\
    RN~\cite{RN}&99.6 $\pm$ 0.2\%&99.8 $\pm$ 0.1\%&97.6 $\pm$ 0.2\%&99.1 $\pm$ 0.1\% \\
    Meta-GAN~\cite{MetaGAN}&99.67 $\pm$ 0.18\%&99.86 $\pm$ 0.11\%&97.64 $\pm$ 0.17\%&99.21 $\pm$ 0.1\% \\
    \midrule
    PARN(ours)&\textbf{99.91 $\pm$ 0.08\%}&\textbf{99.93 $\pm$ 0.03\%}&\textbf{98.55 $\pm$ 0.18\%}&\textbf{99.48 $\pm$ 0.05\%} \\
    \bottomrule
    \end{tabular} }
    \caption{Few-shot classification accuracies on Omniglot. Results are mean accuracies over 1000 test episodes with 95\% confidence intervals. `-': not reported}
    \label{table4}
\end{table*}

\begin{table}[tb]
    \small
    \centering
    \setlength{\tabcolsep}{1.4mm}{
        \begin{threeparttable}
            \begin{tabular}{ccc}
            \toprule
            \multirow{2}{*}{Method} &\multicolumn{2}{c}{\textbf{5-way Acc.}}\\
            \cmidrule(l){2-3}
            &\begin{tabular}[c]{@{}c@{}}1-shot\end{tabular}&\begin{tabular}[c]{@{}c@{}}5-shot\end{tabular} \\
            \midrule

            Meta-LSTM~\cite{MetaLSTM}&43.44 $\pm$ 0.77\%&60.60 $\pm$ 0.71\% \\
            MAML~\cite{MAML}&48.70 $\pm$ 1.84\%&63.11 $\pm$ 0.92\% \\
            Meta-GAN~\cite{MetaGAN}&52.71 $\pm$ 0.64\%&68.63 $\pm$ 0.67\% \\
            \midrule
            MMNets~\cite{MMNets}&53.37 $\pm$ 0.48\%&66.97 $\pm$ 0.35\% \\
            \midrule
            Matching Nets~\cite{MatchingNets}&43.40 $\pm$ 0.78\%&51.09 $\pm$ 0.71\% \\
            Matching Nets FCE~\cite{MatchingNets}&43.56 $\pm$ 0.84\%&55.31 $\pm$ 0.73\% \\
            Proto Nets~\cite{ProtoNets}\tnote{1}&44.53 $\pm$ 0.76\%&65.77 $\pm$ 0.70\% \\
            Proto Nets~\cite{ProtoNets}\tnote{2}&49.42 $\pm$ 0.78\%&68.20 $\pm$ 0.66\% \\
            RN~\cite{RN}&50.44 $\pm$ 0.82\%&65.32 $\pm$ 0.70\% \\
            \midrule
            RN\tnote{3}&51.64 $\pm$ 0.83\%&66.08 $\pm$ 0.69\% \\
            PARN(ours)&\textbf{55.22 $\pm$ 0.84\%}&\textbf{71.55 $\pm$ 0.66\%} \\
            \bottomrule
            \end{tabular}
            \begin{tablenotes}
                \footnotesize
                \item[1]Trained with 5-way 15 queries per episode task, which is the same as us.
                \item[2]Trained with 30-way 15 queries per episode task.
                \item[3]Our reimplementation of RN~\cite{RN}.
            \end{tablenotes}
        \end{threeparttable}}
    \caption{Few-shot classification accuracies on Mini-Imagenet. Results are mean accuracies over 600 test episodes with 95\% confidence intervals.}
    \label{table3}
\end{table}

Besides, as mentioned in Section~\ref{Sec1}, we propose DFE to handle the situation (i) where two objects are in different positions, and DCA to deal with the situation (ii) where related fine-grained features are in different positions. Comparing the results of DFE in Table~\ref{table1} and DCA in Table~\ref{table2}, we can find that DCA contributes much more than DFE. According to our analysis, one reason is that in datasets the situation (ii) occurs more commonly than the situation (i), so the effect of DCA can be more apparent. Another reason is that since DCA can compare related features in any position, it naturally has a certain ability to deal with the situation (i). In other words, DCA is general for the two situations.

\subsection{Comparison with the State-of-the-arts}

In this subsection, we combine DFE and RN with DCA as our proposed position-aware relation network (PARN) to compare with previous state-of-the-art approaches on Mini-Imagenet and Omniglot.

\textbf{Mini-Imagenet}\quad The results on Mini-Imagenet are summarized in Table~\ref{table3}. The first three methods in Table~\ref{table3} are optimization-based, and the fourth method (MMNets) is memory-based. Others methods, including ours, are metric-based. The result of our reimplementation of RN~\cite{RN} is better than the reported because our $2\times 2$ max-pooling layers are applied in the last two layers but not the first two, and avoid premature loss of information. Compared with the optimization-based~\cite{MAML,MetaLSTM,MetaGAN} and memory-based methods~\cite{MMNets}, our proposed PARN achieves better accuracies without the need for updating the model for new tasks or introducing complicated memory structure. As for metric-based methods, after combining DFE and DCA, PARN improves RN from 51.64\% to 55.22\% in the 1-shot task and 66.08\% to 71.55\% in the 5-shot task, and defeats all the other metric-based methods by a clear margin. In summary, our proposed method achieves state-of-the-art performance.

\textbf{Omniglot}\quad The experimental results on Omniglot are shown in Table~\ref{table4}. Most previous methods have performed quite well on the Omniglot dataset. However, in all 1-shot and 5-shot tasks, our method still outperforms them by a comparable margin and reaches state-of-the-art results. It is worthy to notice that our 5-way 1-shot result even outperforms the previous 5-way 5-shot results.

\section{Conclusion}

In this paper, we propose the position-aware relation network (PARN), a more effective and robust deep metric network for few-shot learning. Firstly, we introduce the deformable feature extractor (DFE) to extract more efficient features, which is beneficial for the subsequent comparison efficiency and precision. Secondly, by introducing only a small number of parameters, our proposed dual correlation attention mechanism (DCA) helps RN overcome its inherent local connectivity to compare related semantic objects or fine-grained features in different positions. Therefore, our model is more flexible and robust to learn metrics. Last but not least, we validate our proposed approach on Omniglot and Mini-Imagenet, which achieves state-of-the-art performance.

\section{Acknowledgments}

This work is supported in part by the Guangzhou Science and Technology Program key projects (No. 201707010141, 201704020134), GD-NSF (no.2017A030312006), the National Natural Science Foundation of China (Grant No.: 61771201), the Program for Guangdong Introducing Innovative and Enterpreneurial Teams (Grant No.: 2017ZT07X183).

{\small
\bibliographystyle{ieee_fullname}
\bibliography{arxiv_submission}
}

\end{document}